# Towards a constructive multilayer perceptron for regression task using non-parametric clustering. A case study of Photo-Z redshift reconstruction

C. Arouri  E. Mephu Nguifo  S. Aridhi  C. Roucelle  G. Bonnet-Loosli  N.Tsopzé
Department of Computer Science LIMOS, BP 10448, F-63000 Clermont-Ferrand, France

**Abstract**

The choice of architecture of artificial neuron network (ANN) is still a challenging task that users face every time. It greatly affects the accuracy of the built network. In fact there is no optimal method that is applicable to various implementations at the same time. In this paper we propose a method to construct ANN based on clustering, that resolves the problems of random and ad'hoc approaches for multilayer ANN architecture. Our method can be applied to regression problems. Experimental results obtained with different datasets, reveals the efficiency of our method.

Keywords: regression, ANN architecture, accuracy.

## 1. Introduction

Artificial neural network (ANN) is a model of information processing schematically inspired by biological neurons. The key of this paradigm is its structure composed of a large number of processing elements strongly connected (neurons), able to perform in groups and in parallel to solve specific problems connected by adaptive synaptic connections whose values are connected through a learning algorithm. Although the link with biology has been a major motivation for the pioneers of artificial intelligence, the real development of ANN is purely mathematical and statistical information which made the ANN required in various areas among which imaging, pattern recognition, predicting weather, classification of radar disasters. They have been applied in problems of signal processing and time series, and also in optimization problems such as task scheduling and multi-level tasks allocation. Artificial neural networks are learning techniques that have been widely used to solve regression and classification problems. Yet anyone looking to use a multilayer artificial neural network in a given application is facing a difficult choice, which is the choice of architecture, what is the number of hidden layers and how to select the number of neuron in each layer? The question that arises just after holding the components of the network is, how to know that this choice is reliable? If it is not, how do we know that? Wrong choice may lead to drawbacks in the quality of learning and performance of the network built. Many questions that researchers have tried to answer, but the task remains difficult. There is no optimal method for MLP (MultiLayer Perceptron) architecture which is applicable in various applications at a time. At the beginning and in the absence of any preferred option, users try several architectures randomly and then choose the one that gives the best result, but this is not practical when dealing with huge amounts of data. Subsequently a variety of methods have emerged, we mention: the ad'hoc method, static methods among them those using prior knowledge about the learning as KBANN [1], as well as other static constructive methods based on concept lattice or deduced from a decision tree and finally dynamic approaches where neurons are created at the same time as learning without generating random weights in advance.

Among the dynamic approaches, stands out the Distal (Inter-pattern distance-based constructive learning algorithm) method [2], based on a 3-layer architecture, with a single hidden layer, Distal [2] built the ANN, by grouping training data in regions bounded by thresholds without using a clustering algorithm. Distal is based on the distance between the observations and their classes. Unfortunately this method is appropriate only to build a 3-layered perceptron for classification task, and can't be applied in the case of a regression.

As most constructive methods of MLP are not applicable in the case of a regression, we try to exploit a new way towards a constructive multilayer perceptron for regression task based on a clustering algorithm. The idea is to make a correspondence between the number of clusters obtained from data and the number of neurons in the hidden layer, and then use quasi-Newton learning algorithm to train the network.

In this paper we present an extension of DisAl algorithm to handle regression tasks, we deal with this problem within a static manner. We demonstrate the applicability of this method on some practical problems as photometric redshift. The remainder of this paper is organized as follows: Section II presents the methodology of our method and describe the steps of the algorithm. Sections III gives an overview of data, and IV illustrates the results obtained. Section V concludes with a discussion and some directions for future research.

## 2. Methodology

Most constructive neural network learning algorithms are limited to classification problems or require prior knowledge. They are not adapted to regression problems. In order to find a new solution to choose the architecture of ANN and that can be applied in the case of a regression to approximate real values, we proposed a method that generalize other constructive approaches like DistAl [2] to deal with regression tasks. The method requires the use of a nonparametric clustering algorithm that produces an optimal number of clusters without specifying the number of clusters in advance. The number of clusters obtained is used to fix the number of neurons in the hidden layer.

### 2.1 Non-parametric clustering

The nonparametric clustering algorithms don't need to provide the number of clusters in advance, indeed it overcomes the problems posed by some parametric algorithms, such as Kmeans, that require setting the number of clusters k before starting the data partitioning algorithm. This number k, set in advance, may not be the right number which leads to improper distribution of the data. So instead of providing the number of clusters k in advance, the idea is to use a clustering algorithm that searches for the optimal number of clusters. Among these non-parametric algorithms, we are interested in Xmeans [3], DBscan [4] and Meanshift [5].

**Xmeans:** It is based on K-means and requires the specification of a search space of possible values of the optimal number of clusters. Xmeans proceeds by successive application of 2-means and needs the comparison criterion BIC [6]. The following steps summarize the idea of X-means:

1. The X-means algorithm starts by applying k0-means the set S to obtain C1, C2... Ck0 clusters.
2. Repeat, for each cluster, apply 2-means then choose the 2 new centroids in opposite directions, and proportionally to the the size of the parent cluster.
3. Compare the BIC of parent cluster to the BIC of son, if it is better then switch to another cluster; otherwise replace the father by these two sons.
4. If the higher bound of the search space is reached or the maximum number iteration is reached then stop else return 2.

**DBscan:** The idea of the algorithm is for any observation point, not yet assigned to a cluster, retrieve its neighborhood at a distance $\xi$, compare the number of neighbors to MinPts, if it is a noise then use another point, otherwise its neighbors course step by step the same way.

**Meanshift:** It is an iterative and powerful algorithm, applied especially in image segmentation, it represents the space of descriptors (attributes) by a probability density function, this density is determined by the density estimator kernel. The mean shift procedure is repeated for each item, calculate the neighborhood in a window, calculate the meanshift and shift this item to meanshift. Meanshift converges to the zeros of the gradient, which are the local maxima of the estimator. The set of points that converge to the same point of attraction (local maximum) represents a cluster.

### 2.2 Neural Networks

Multilayer perceptron (MLP) is a static neural structure composed of successive layers which communicate and exchange information through synaptic connections represented by an adaptative weight. The structure of multilayer network contains an input layer which is made of number of perceptions equal to the number of data attributes, on the other hand, output layer includes one perceptron in the case of regression or more when it is a task of classification and in this case the number of perceptron is equal to the number of classes to predict, all the other layers are considered as hidden.

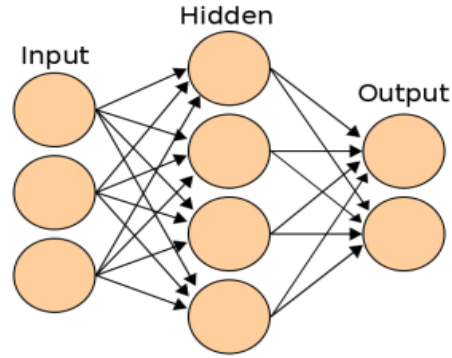

Figure1: Neural networks with 3couches, 3 inputs, 2 outputs and 4 hidden nodes

MLPQNA is a learning rule algorithm, it's an optimization the Newton method ,it seeks local maxima and minima and includes several techniques. Among them stands out the technique of Broyden-Fletcher-Goldfarb-Shanno [7] also known as BFGS algorithm tested by Watrous and compared to the backpropagation algorithm and Fletcher algorithm Powell [8], it estimates the inverse of the symmetric positive definite the Hessian matrix of the cost function in $O(n_2)$ compared with the method of Newton $O(n_3)$, without having to calculate it automatically. It nevertheless poses some problems when approximating the Hessian if it is large N * N, as it needs the storage of the whole matrix. Thus the algorithm L-BFGS, BFGS optimization, based on the idea of storing in memory only some significant vectors of the Hessian and not its entirety.

### 2.3 Construction of Neural Networks

The idea of the method is detailed in figure1 can be summarized in three steps:

1. The first step is a phase of clustering where data is shared in different groups, using a non-parametric clustering algorithm.
2. The definition of the architecture, as in the method Distal ANN contains 3 layers: an input, a single hidden layer and a single output node as we are studying a case of regression. The number of nodes of the hidden layer is the number of clusters returned by the non-parametric clustering algorithm.
3. In the third and final step, the ANN is trained with a learning algorithm, such as MLPQNA algorithm (Multi layers Perceptron Quasi-Newton algorithm) [9].

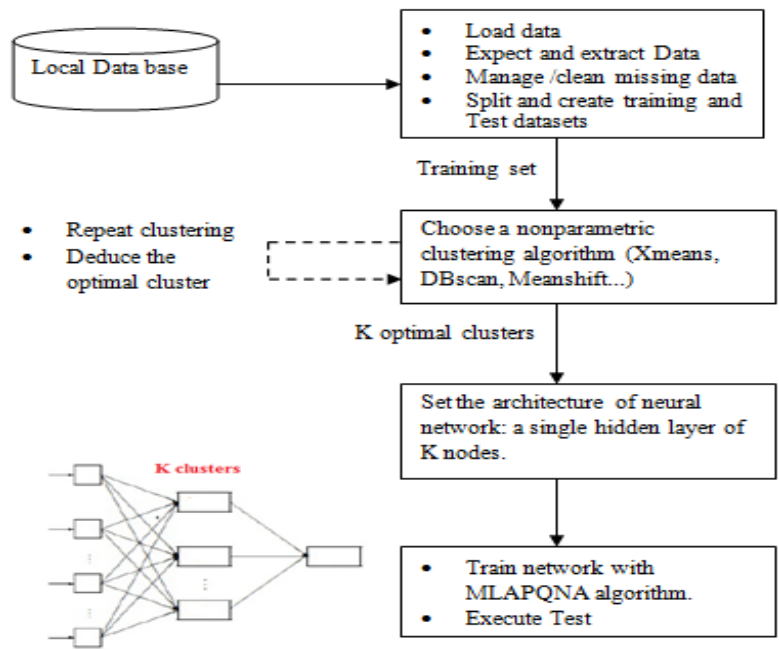

Figure1: Different steps of our method

## 3. Photometric redshift

The expansion of the universe causes the spacing of galaxies that are gravitationally bound. During this expansion, we get a modified galaxies radiation, their spectra are shifted to the red. Two categories of methods have been developed to measure the photometric redshift: Template fitting method that adjust SED (spectral energy distribution) and require the construction of models for comparison. Empirical methods use spectroscopic redshift to calibrate the learning algorithm and overcome the problems of uncertainty within models built in template fitting method.

**PHAT:**
The PHAT (PHoto-z Accuracy Testing) project [10] provides a standard test environment to estimate redshift and compare the obtained results. We distinguish PHAT1 a real data catalog for redshift estimation, which includes 1984 objects where only the fourth spectroscopic values are used for training because all other values are set to z-spec = -9.999, so, we need an extraction of useful data for training.

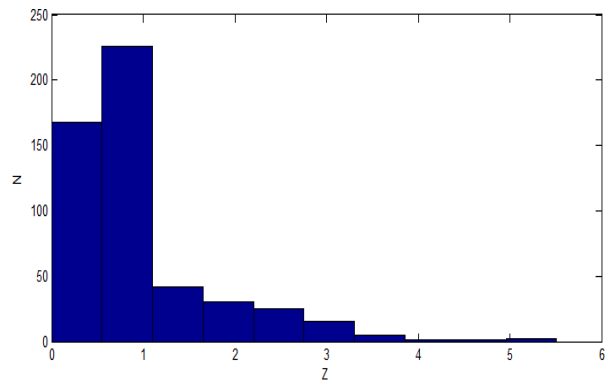

**Figure3: Histogram of the data (515 galaxies)**

At the attribute level, PHAT1 includes 18 attributes, among them there are irrelevant values at 99 and -99, and these values are due to sampling errors and may bias the step of training. We note also from Figure2 that 316 recordings from 515 retained for training, don't contain wrong values, that's why we test these two data sets separately, thus we kept the one that gave us the best and worst value of RMS (Root Mean Square) error.

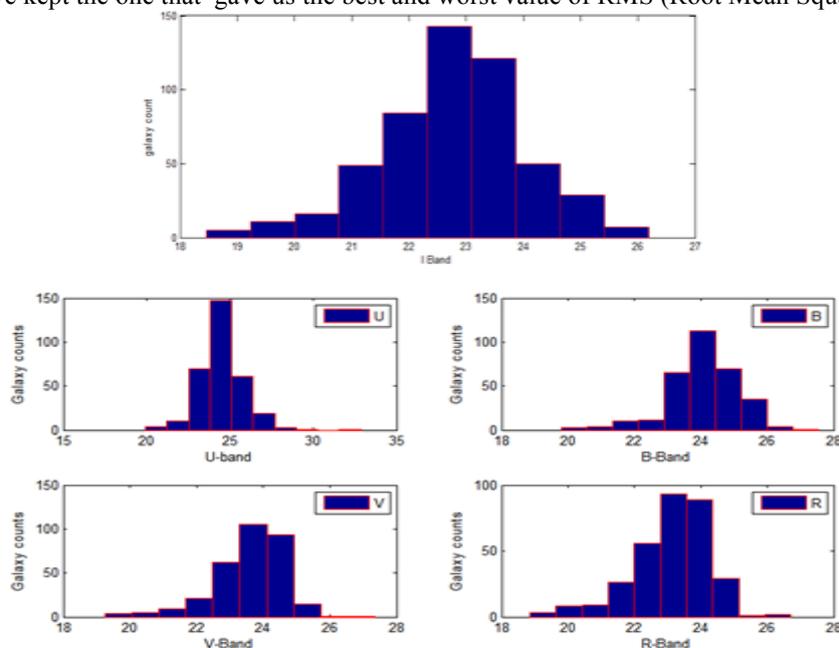

**Figure4 Distribution of the bands I, U, B, V and R**

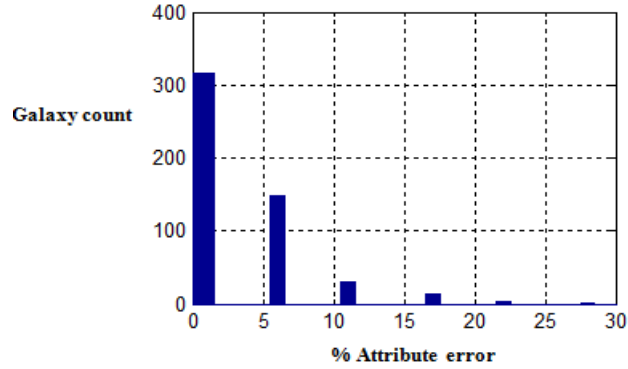

**Figure2: Histogram of the data according to the percentage of error in each attribute.**

## Deep2 D4:

Deep2 uses multi-object spectrograph DEIMOS and observes four regions with a different telescope KECK2 10m. Deeep2 D4 catalog is the fourth updated version, it contains 50319 redshift galaxies ranging from 0 to 1.4. It also includes many features of galaxies such as: name, position in two dimensions over the spectral data.

## 4. Experimentation

In this section we discuss the results of our method. For this, we used different datasets as PHAT1 [10] or Deep2 D4 [11] to estimate redshift and a set of ordinary real data. To properly carry out the experiments, we used two reported and available toolbox: ANNZ [12] and Skynet[13], where the choice of multilayer architecture remains a difficult task.

### 4.1 PHAT1 data

Whatever the data used, they are split within a holdout into two sets, in order to keep 70% for training and 30% for the test. As we mentioned in the introduction, the most naïve method to construct an MLP, is to select randomly several architecture then choose the description of ANN that provides minimal error and a better quality of learning. That's exactly what we've done in Figure 4. To perform a regression we use the set that contain 316 objects, after preparing the training and test set we use a network that contains a single hidden layer and we vary K the number of hidden nodes in the unique hidden layer. The best and more adequate number of nodes is the one for which the error is minimal and the correlation [13] is maximal. Thus, According to Figure 4 we note that when the error decreases the correlation increases and we deduce that 9 is the best number of hidden neurons giving a minimum error, with error as close to 8,10 and 11 neurons. This approach is not applicable in the case of big amounts of data, since it is time consuming. For this we looked through a nonparametric clustering algorithm to verify this result.

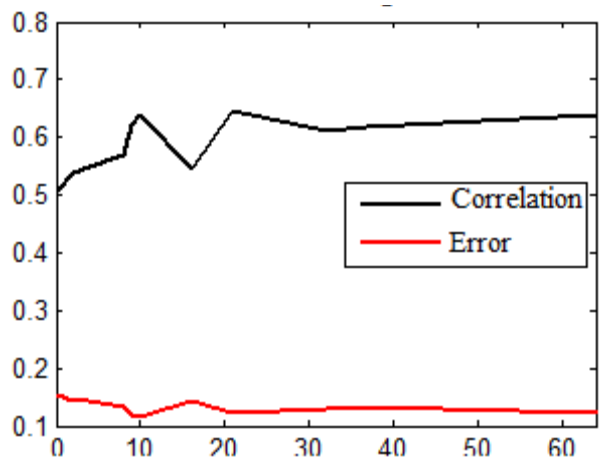

**Figure4: Effect of varying the number of hidden neurons**

To check the robustness of our method, we launch Xmean on the cleaned dataset composed of 316 galaxies and we obtain 9 clusters, the performance of the network built with 9 hidden neurons is evaluated in table2. By applying Xmeans on 70% of 515 galaxies (the set that contains the most error attributes) and varying several input values kmin (lower bound of the search space) of Xmeans, we obtain the optimal number of clusters k for each configuration, and thus recover time clustering. Then we train an ANN that contains one single hidden layer of neurons with quasi Newton algorithm. Table 1 summarizes the obtained results and shows that despite the variation of the initial configuration, the number of clusters of results returned by X-means closed to 21 clusters. We also conclude that the clustering time is negligible compared to the training time. By detailing the related statistics with a single layer composed at first of 9 neurons, then of 21 hidden neurons in table 2, we noticed the efficiency of our method.

The accuracy of our method tested on PHAT1 is good compared to the neural network-based methods that have been tested with PHAT1 and to the optimal one obtained with experiments performed randomly and in an ad 'hoc manner. RMS values and biases in table2 are calculated without removing the outliers.

**Table 1 . Comparison of training time and clustering time**

| Kmin | Number of clusters | Clustering time (sec) | ANN training time (sec) |
|---|---|---|---|
| 2 | 20 | 0.29 | 20 |
| 4 | 21 | 1.18 | 19 |
| 6 | 21 | 1.18 | 19 |
| 7 | 21 | 1.38 | 19 |
| 8 | 25 | 1.58 | 22 |
| 10 | 21 | 1.58 | 19 |

We conclude from the results of experiments on two datasets PHAT1, that if a nonparametric clustering algorithm is held once, it is enough to infer the architecture of ANN which provides a good quality of network in training and test sets. Moreover as time clustering is still negligible compared to the training time, so we can remedy the naive and ad hoc method and repeat clustering several times in order to guarantee the stability of the number of clusters.

Table2. Evaluation of ANN's architecture 18:9:1 and 18:21:1

| Architecture | Norm.RMS | Bias | Outliers>0.15 |
|---|---|---|---|
| 18:9:1 on 316 galaxies | 0.071 | 0.0745 | 31.57% |
| 18:21:1 on 515 galaxies | 0.0747 | 0.0074 | 30.96% |

We have also tested the catalog deep2 D4 [11] and compliant concrete_data from UCI (machine learning repository). At first we decided for both deep2 and concrete_data to test network that contains one single hidden layer which includes neurons selected randomly or using ad 'hoc method. After that we test our method such as the number of neurons in the hidden layer is justified by a clustering algorithm. In these experimentations we use Xmeans, so we vary the space search borned by $k_{min}$, in order to infer a better clustering result.
Table3 shows two types of data, on each one we proceed two kinds of test. In the first test, it comes to change the number of neuron randomly or with an ad'hoc manner. In the second test we deduced the number of neurons in the hidden layer from a nonparametric clustering algorithm.
We note that network built with a clustering algorithm is more efficient and generates more accurate output than randomly selected one. It comes that RMS values in the two test are close enough but the clustering time is negligible compared to the training time and this shows the robustness of the method.

Table3 Evaluation of ANN's architecture built with and without clustering

| | Hidden nodes | N°clusters | Clustering time (min:sec) | RMS validation | RMS test | Training time (min:sec) |
|---|---|---|---|---|---|---|
| Deep2 without clustering | 2 | - | - | 0.75021 | 0.7535 | 23:25 |
| | 4 | - | - | 0.7475 | 0.7514 | 26:17 |
| | 7 | - | - | 0.748 | 0.75268 | 17:16 |
| Deep2 with clustering | **kmin** | **N°clusters** | **Clustering time** | **RMS validation** | **RMS test** | **Training time** |
| | 2 | 108 | 00:10 | 0.75021 | 0.7535 | 23:25 |
| | 4 | 88 | 00:09 | 0.7475 | 0.7514 | 26:17 |
| | 6 | 91 | 00:10 | 0.748 | 0.75268 | 17:16 |
| | 40 | 91 | 00:10 | 0.748 | 0.7526 | 17:16 |
| Concrete_data without clustering | **Hidden nodes** | **N°clusters** | **Clustering time** | **RMS train** | **RMS test** | **Training time** |
| | 2 | - | - | 0.25 | 0.25 | 00:03 |
| | 7 | - | - | 0.17 | 0.20 | 00:03 |
| | 8 | - | - | 0.1584 | 0.1884 | 00:04 |
| Concrete_data with clustering | **kmin** | **N°clusters** | **Clustering time** | **RMS train** | **RMS test** | **Training time** |
| | 7 | 10 | 00 :00.15 | 0.1477 | 0.1884 | 00:03 |

## 4 Conclusion and remarks

The method described here extends the constructive MLP DistAl method, and allows dealing with regression tasks. This method relies on non-parametric clustering to fix the 3-layer network, and thus avoids the user to provide this parameter. Experimentations shows that the constructing method of ANN with a first step of clustering is adequate for applications that process large amounts of data, in fact we find that the clustering time is still negligible compared to the ANN training time, thus running a clustering algorithm to determine the number of hidden neurons is preferable to the repetitive process of training neural network to find the optimal architecture. Besides the method gives good results compared to the ad'hoc methods where a repetitive process is run to find the optimal architecture. The results of the method presented in this paper correspond to those performed with Xmeans algorithm, but others algorithms can also be used, such as DBSCAN or Meanshift.

**Acknowledgment**
This work was supported by the PETASKY project for Astrophysics research.